\documentclass{article}

\usepackage[final]{neurips_2019}

\usepackage[utf8]{inputenc}
\usepackage[T1]{fontenc}
\usepackage{hyperref}
\usepackage{url}
\usepackage{booktabs}
\usepackage{amsfonts}
\usepackage{nicefrac}
\usepackage{microtype}
\usepackage{graphicx}
\usepackage{xcolor}
\usepackage{lipsum}
\usepackage{amsmath}
\usepackage{multirow}
\usepackage{graphicx}

\title{
Integrating Domain Knowledge for Financial QA: A Multi-Retriever RAG Approach with LLMs \\
}

\author{
  Yukun Zhang \\
  Department of Computer Science \\
  Stanford University \\
  \texttt{yukzhang@stanford.edu} \\
   \And
   Stefan Elbl Droguett \\
  Graduate School of Business \\
  Stanford University \\
  \texttt{selbl@stanford.edu} \\
  \And
   Samyak Jain \\
  Graduate School of Business \\
  Stanford University \\
  \texttt{samyakj@stanford.edu} \\
}

\begin{document}

\maketitle

\begin{abstract}

This research project addresses the errors of financial numerical reasoning Question Answering (QA) tasks due to the lack of domain knowledge in finance. Despite recent advances in Large Language Models (LLMs), financial numerical questions remain challenging because they require specific domain knowledge in finance and complex multi-step numeric reasoning. We implement a multi-retriever Retrieval Augmented Generators (RAG) system to retrieve both external domain knowledge and internal question contexts, and utilize the latest LLM to tackle these tasks. Through comprehensive ablation experiments and error analysis, we find that domain-specific training with the SecBERT encoder significantly contributes to our best neural symbolic model surpassing the FinQA paper's top model, which serves as our baseline. This suggests the potential superior performance of domain-specific training. Furthermore, our best prompt-based LLM generator achieves the state-of-the-art (SOTA) performance with significant improvement (>7\%), yet it is still below the human expert performance. This study highlights the trade-off between hallucinations loss and external knowledge gains in smaller models and few-shot examples. For larger models, the gains from external facts typically outweigh the hallucination loss. Finally, our findings confirm the enhanced numerical reasoning capabilities of the latest LLM, optimized for few-shot learning.

\end{abstract}

\section{Introduction}
%The introduction explains the problem, why it's difficult, interesting, or important, how and why current methods succeed/fail at the problem, and explains the key ideas of your approach and results. Though an introduction covers similar material as an abstract, the introduction gives more space for motivation, detail, references to existing work, and to capture the reader's interest.

\begin{figure}[h]
    \begin{center}
        \includegraphics[width=0.8\textwidth]{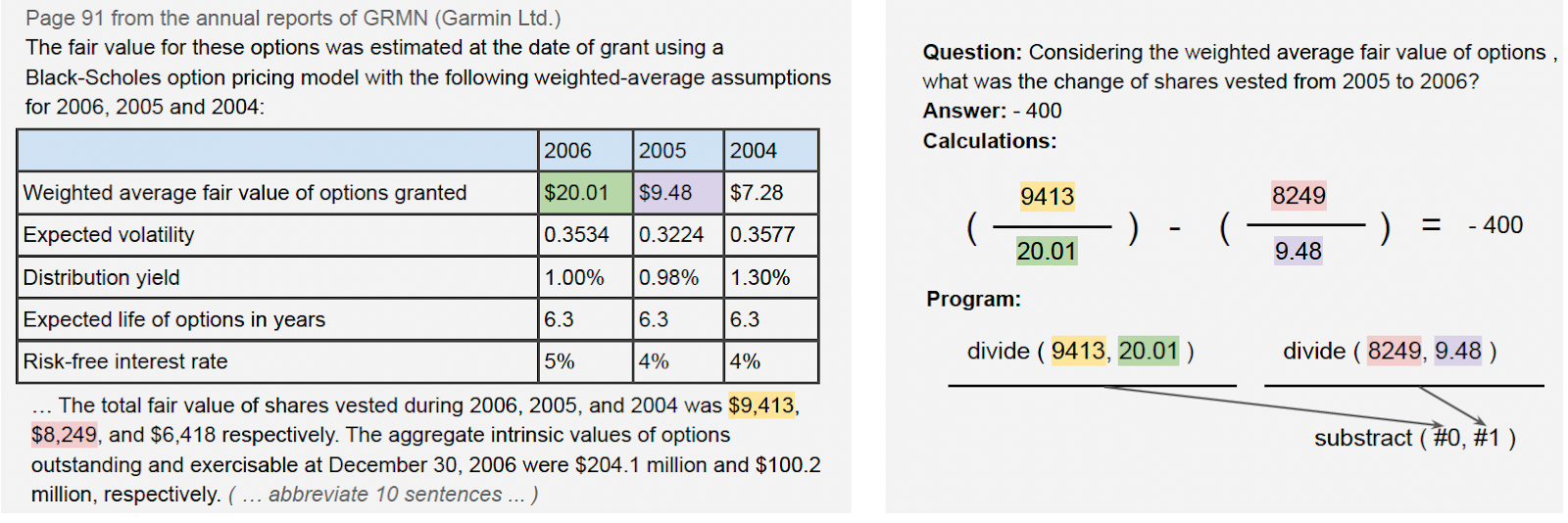}
        \caption{Sample question and answer from the FinQA dataset}
        \label{fig:question}
    \end{center}
\end{figure}

Financial numerical reasoning QA tasks present a challenge for current LLMs as they require an understanding of dense financial terminology and complex computation to reach the correct answer. We attempt to mitigate this issue by leveraging additional external context information and the capabilities of LLMs. For example, a model may not understand financial terminologies such as 'options' and 'fair value,' which are not explained in the query contexts (see Figure~\ref{fig:question}). We propose that access to external financial dictionaries (like Investopedia) to look up financial definitions as non-parametric memory can mitigate this issue. Inspired by RAG~\cite{lewis2020rag}, we aim to build a powerful multi-retriever RAG system with two options for the generator (neural symbolic and LLM) to better retrieve key facts from both internal query contexts and external terminology explanations to generate final answers with higher accuracy. 

\section{Related Work}

Existing models for financial numerical reasoning QA tasks still fall short of human expert performance. A recent survey~\cite{lee2024survey} on finance LLMs finds that the current SOTA model for financial QA tasks is the zero-shot GPT-4 with 75\% exact match accuracy, significantly below the 91\% accuracy of human experts. According to the FinQA paper~\cite{chen-etal-2021-finqa}, over 85\% of the errors stem from: (i) lack of domain knowledge in finance; or (ii) numerical reasoning errors. To address the latter, the ConvFinQA paper~\cite{chen-etal-2022-convfinqa} adopted a method similar to Chain of Thought~\cite{wei2022CoT} on a financial QA dialogue dataset and showed a significant improvement for the outputs.

Inspired by the RAG model, which accesses external knowledge without additional training and outperforms SOTA results in many NLP tasks, we propose incorporating finance-specific non-parametric memory to help the model understand domain-specific concepts and improve performance. We follow the approach in the DPR (Dense Passage Retrieval) - FAISS paper~\cite{karpukhin-etal-2020-dense} to design our external retriever, which combines DPR with efficient similarity search for superior performance.

\section{Approach}

\begin{figure}[htbp]
    \centering
    \includegraphics[width=\textwidth]{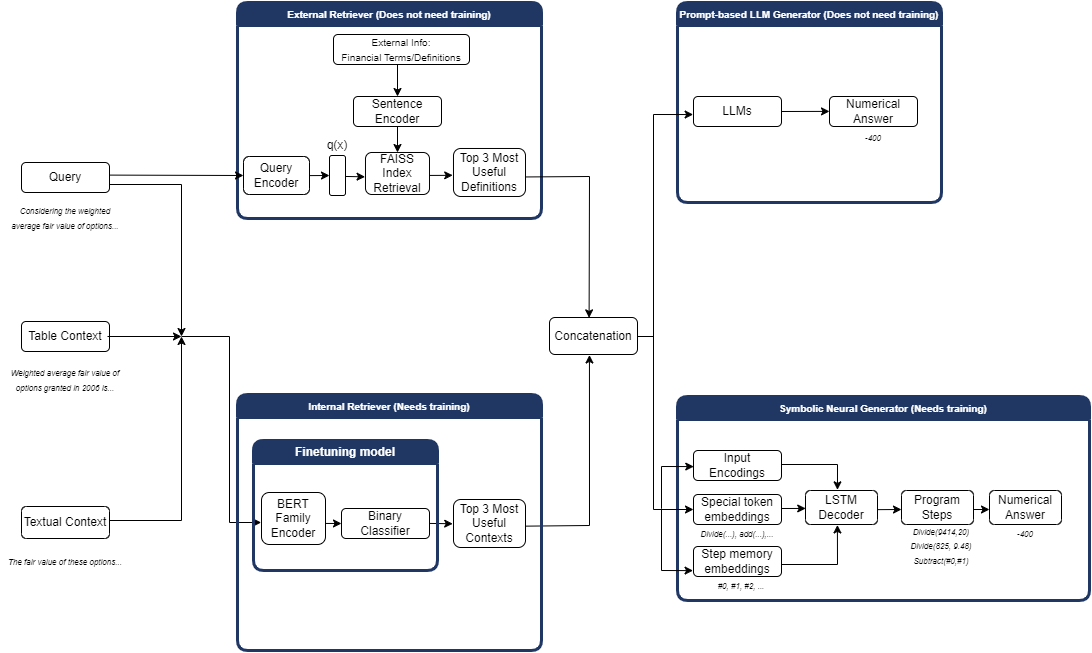}
    \caption{Model's architecture}
    \label{fig:architecture}
\end{figure}

Figure~\ref{fig:architecture} depicts our model's architecture. Given that the query context can be long (over 2600 tokens) and contains irrelevant information, we focus on retrieving only the useful facts to improve system efficiency. The architecture includes two retrievers (internal and external) and a generator with two options for generating the final answers: a prompt-based LLM generator and a symbolic neural generator. We describe each part in detail below:

\textbf{Internal Retriever}: We fine-tune BERT-family models~\cite{devlin2019bert} to train a binary classifier for extracting the most relevant in-context supporting facts. We follow Hugging Face's steps to fine-tune pretrained models. Adopting the MathQAExample class from FinQA example code, we create question-context sentence pairs with appropriate padding and tokenization. Each sentence is labeled as positive or negative based on pre-labeled golden fact sentences. We use the Adam optimizer and CrossEntropyLoss for binary classification, and we add gradient clipping and a ReduceLROnPlateau learning rate scheduler to ensure consistent training loss reduction. For each query, we select the top 5 and top 3 sentences from the model outputs based on logit score rankings.

\textbf{External Retriever} We use the DPR-FAISS structure similar to RAG to extract external supporting facts most relevant to the query from outsourced data without additional training. We begin by using an encoder to generate embeddings for each one-sentence definition summary in the financial terminology dictionary. Since FAISS is based on inner product similarity, we apply L2 normalization to the embeddings. We then use FAISS for fast retrieval, retrieving the top 3 related definitions for each query and storing them as external domain knowledge with associated similarity scores. 

\textbf{Prompt-based Generator}: To fit in our system, the prompt-based LLM generator needs to be an encoder-decoder model. Using a paid API in Google Colab, we leverage Gemini Pro models~\cite{reid2024gemini} as our main LLM generator and feed the outputs from the retrievers as inputs to the LLM. Given an explicitly specified instruction prompt (see Appendix A for details) concatenated after the query, the model generates the final answer as a number or yes/no, whichever is applicable. Zero-shot and few-shot examples (see Appendix A for prompting details) are also tested on the dev set.

\textbf{Symbolic Neural Generator}: This generator requires training and consists of an encoder and decoder. We set up the same operation tokens (such as “add(” ), constant tokens (such as “const\_100”) and step memory tokens (such as \#0, \#1) defined in the FinQA paper to formulate the program steps. The operation tokens are the 10 most common types of math operations used to answer our queries; the constant tokens represent common constant numbers used in financial calculations and the step memory tokens denote the result from the $n$-th step (see Appendix B for details). We combine operation tokens and constant tokens, referred to as "special tokens", in the model diagram.

Each program step token can come from either the numbers in the retrieved contexts or from the three types of tokens described above. Our generator aims to generate each program step token one at a time based on its index (all numbers in the retrieved contexts and the three types of tokens are stored in the "program\_ids" variable). For example, the single program step token "subtract(" has the index "5," and if the output of our model is "5," it will be converted to the string "subtract(".

To begin with, we randomly initialize the embeddings of these three token indexes as $h_i^o$, $h_i^c$, $h_i^s$. Then we use pretrained BERT family models to encode the retriever output as $h_i^i$. Our LSTM decoder receives the entire token embeddings as: $$H = [h_i^i, h_i^o,  h_i^c, h_i^s]$$

We use the decoder output $h_T$ to calculate two attention vectors, $att_i$ and $att_h$, for input sequence attention and decoder history attention, respectively. We define a context vector $c_T$ to combine all the contextual information from input sequences and the decoder history, where: $$c_T = W_c[att_i; att_h; h_T]$$

A third attention vector $att_r$ for input sequence is also added for the reasoning path of the program, so the entire reasoning results are: $$H_T = W_h[H; H \circ att_r]$$

The final prediction output of the model is generated by choosing the top program index from the softmax results: $$w_T = \text{softmax}(H_T \cdot c_T)$$

We use the Adam optimizer and CrossEntropyLoss for training. Additionally, we incorporate gradient clipping and a ReduceLROnPlateau learning rate scheduler to ensure the training loss consistently decreases over time. During inference, we use masks at each decoding step to ensure the structural correctness of the generated programs. The direct outputs of the program step tokens from the generator are concatenated to form the complete program step string. For instance, the golden generated program step for the example query in Figure 1 is the string “divide(9413, 20.01), divide(8249, 9.48), subtract(\#0, \#1)”. We adopt the functions from the FinQA paper to evaluate program steps and compute the final numerical answers from the given program steps. The final answer is the number -400.

\textbf{Baseline:} using our new system, we re-implement the best-performing FinQANet structure from the original FinQA paper, training for 20 epochs. This single retriever-generator architecture employs a BERT-base-encoded retriever for internal context and a RoBERTa-large-encoded~\cite{liu2019roberta}, LSTM-decoded~\cite{hochreiter1997long} generator. The results are shown in the later experiment section.

\section{Experiments}
%This section contains the following.

\subsection{Data}
%Describe the dataset(s) you are using (provide references). If it's not already clear, make sure the associated task is clearly described.
%Being precise about the exact form of the input and output can be very useful for readers attempting to understand your work, especially if you've defined your own task.

For our internal QA data, we use the original FinQA dataset. Based on the earnings reports of S\&P 500 companies, FinQA is an expert-annotated dataset containing 8,281 financial QA pairs, along with labeled numerical reasoning processes and highlighted supporting sentences. It is released as training (6,251), dev (883), and test (1,147) sets. Figure~\ref{fig:question} shows an example question. The data stores query question texts, the textual contexts, and tabular contexts in JSON format. The average number of tokens per question is 687, and the maximum is 2,679. We improved the pre-processing methods from the FinQA analysis to transform each row of the table into long sentences based on various table formats. For example, in Figure~\ref{fig:question}, the first entry of the last table row is rewritten as "Risk-free interest rate of 2006 is 5\%." This ensures that all inputs are in sentence textual formats.

For our external domain knowledge data, we use the FinRAD dataset~\cite{ghosh-EtAl-2022FNP-FinRAD}, which includes explanations and definitions for over 13,000 financial terms and words. We pre-processed the data by replacing and removing unusual symbols and correcting typos. Although most definitions are one or two sentences long with an average of 46 tokens, we find many long definitions with examples and formulas, with a maximum token length of 667. To address this, we leveraged Gemini-1.0-pro to create one-sentence summaries for each term definition, and the generated outputs are very precise and consistent based on random human assessments.

\subsection{Evaluation method}

For the supervised internal retriever with fine-tuning, given that our labeled dataset is largely imbalanced (most sentences in the context are irrelevant to the query and are labeled as negative), we use recall to measure the performance for the top 3 and top 5 retrieved facts. For the unsupervised external retriever, we store the cosine similarity scores for the top 3 retrieved definitions.

We evaluate the performance of our generator by measuring execution accuracy (whether the answer is correct) and program accuracy (whether the steps taken by the model to solve the problem match those used by experts when labeling the data). These metrics introduce a trade-off: false positives affect execution accuracy (the model can guess the correct answer), while false negatives affect program accuracy (if the steps taken are not entirely equivalent to the standard solution in the dataset). For the symbolic neural model, we directly calculate the numerical answers using the given program steps to measure execution accuracy. For the prompt-based LLM method, since we do not directly generate the program steps, we add a small $\epsilon$ to allow for some rounding errors in the final outcome. For example, if the model outputs "94.17\%" and the golden answer is 0.942, we consider it correct.

\subsection{Experimental details}

We successfully implement a system with two retrievers (internal and external) and a generator with two options. Due to computational limitations, we could only train each for 1000 epochs. Training time takes over 13 hours for each type of pretrained model in the internal retriever on one RTX 3080Ti GPU and over 8 hours for each type of pretrained model in the symbolic neural generator on one Colab A100 GPU. We set the learning rate to 2e-5, the batch size to 16, and the dropout rate to 0.1.

\textbf{Internal Retriever}: Using the annotated labels from the FinQA dataset, we fine-tune our classifier with five different pretrained models (BERT Base, SecBERT Base~\cite{loukas-etal-2022-finer-sec-bert}, FinBERT Base~\cite{araci2019finbert}, RoBERTa Base and SpanBERT~\cite{joshi2020spanbert}).

\textbf{External Retriever}: Because we do not have labeled data, we utilize five different pretrained models (DPR, BERT Base, SecBERT Base, FinBERT Base and RoBERTa Base) to generate high-quality embeddings for similarity search. We use the FAISS index for fast retrieval to select the top 3 most relevant definitions for each question.

\textbf{Prompt-based LLM Generator}: we leverage the capabilities of current LLMs (Gemini-1.0-pro and Gemini-1.5-pro-latest) and conduct ablation experiments to assess the usefulness of the two retrievers. First, we run zero-shot and few-shot experiments by bundling the contextual texts, table texts, and query questions as inputs to the model. Then, we use the LLM as the single generator with only an internal retriever (BERT-base and SecBERT) and without any external knowledge. Finally, we test the whole system using the LLM as the single generator with both the internal retriever (BERT-base and SecBERT) and the DPR-FAISS external retriever. Additionally, without any fine-tuning, we test both the T5-base~\cite{2020t5} and GPT-2 decoder~\cite{radford2019language} with BERT encoders as a single generator with no retrievers (zero shot), internal retriever only, and both internal and external retriever.

\textbf{Neural Symbolic Generator}: We examine six different pretrained models (BERT Base, BERT Large, SecBERT Base, FinBERT Base, RoBERTa Large, and RoBERTa Base) as the encoder and conduct ablation experiments to assess the effects of the two retrievers, with or without external knowledge. Moreover, we evaluate our model's performance across data subsets for various types of questions (single reasoning step vs. multiple reasoning steps; long context questions vs. short context questions; table-only questions vs. text-only vs. table-text mixed). We use our best-performing neural symbolic model and prompt-based LLMs for the performance breakdown.

\subsection{Results}

\begin{table}[htbp]
    \centering
    \begin{tabular}{lcc}
        \toprule
        \textbf{Fine-tuned Models} & \textbf{Top 3 Recall \%} & \textbf{Top 5 Recall \%} \\
        \midrule
        BERT Base & 88.96 & 92.75 \\
        SecBERT Base & 91.27 & 95.16 \\
        FinBERT Base & 88.94 & 92.83 \\
        RoBERTa Base & 87.76 & 92.67 \\
        SpanBERT & 90.68 & 93.92 \\
        \bottomrule
    \end{tabular}
    \caption{Internal Retriever on Dev Set}
    \label{tab:intRet}
\end{table}

Table~\ref{tab:intRet} shows that for the internal retriever, SecBERT Base gives us much higher recall in both top 3 and top 5 retrieved facts compared to the BERT Base model used in our baseline. Since SecBERT is a BERT model entirely trained on 260,773 publicly available 10-K financial documents, it incorporates much more domain knowledge in finance. This domain-specific training can allow SecBERT to retrieve relevant documents more effectively and reduce the likelihood of missing important information.
%This domain-specific training can allow SecBERT to develop a deeper understanding of financial terminology and context, enabling it to retrieve relevant documents more effectively and reduce the likelihood of missing important information.

\begin{table}[htbp]
    \centering
    \begin{tabular}{lcc}
        \toprule
        \textbf{Sentence Encoders} & \textbf{Median Similarity Score} & \textbf{Mean Similarity Score} \\
        \midrule
        DPR & 0.8583 & 0.8556 \\
        BERT Base & 0.8538 & 0.8527 \\
        SecBERT Base & 0.9287 & 0.9201 \\
        FinBERT Base & 0.8759 & 0.8735 \\
        RoBERTa Base & 0.8744 & 0.8721 \\
        \bottomrule
    \end{tabular}
    \caption{External Retriever on Dev Set for the Top 3 Facts}
    \label{tab:extRet}
\end{table}

Table~\ref{tab:extRet} shows the similarity scores for our external retriever. Due to domain-specific training in the finance domain, we observe the highest similarity scores for the SecBERT encoder.

\begin{table}[htbp]
    \centering
    \begin{tabular}{p{10cm}cc}
        \toprule
        %\textbf{Architecture} & \textbf{Dev Program Accuracy \%} & \textbf{Dev Execution Accuracy \%} \\
\textbf{Architecture (20 Epochs of Training)} & \textbf{Program } & \textbf{Execution} \\
 & \textbf{Accuracy \%} & \textbf{Accuracy \%} \\
        \midrule
        BERT Base Internal Retriever \newline + RoBERTa Large Encoder Generator (Baseline) & 57.87 & 60.02 \\
        \hline
        BERT Base Internal Retriever + DPR-FAISS External Retriever \newline + RoBERTa Large Encoder Generator & 58.49 & 60.96 \\
        \hline
        SecBERT Internal Retriever + RoBERTa Large Encoder Generator & 59.70 & 62.11 \\
        \hline
        \textbf{SecBERT Internal Retriever + DPR-FAISS External Retriever \newline + RoBERTa Large Encoder Generator} & \textbf{60.54} & \textbf{63.48} \\
        \hline
        SecBERT Internal Retriever + RoBERTa Base Encoder Generator & 56.49 & 58.33 \\
        \hline
        SecBERT Internal Retriever + DPR-FAISS External Retriever \newline + RoBERTa Base Encoder Generator & 56.75 & 58.81 \\
        \hline
        SecBERT Internal Retriever + BERT Large Encoder Generator & 53.95 & 55.76 \\
        \hline
        SecBERT Internal Retriever + DPR-FAISS External Retriever \newline + BERT Large Encoder Generator & 54.64 & 56.97 \\
        \hline
        SecBERT Internal Retriever + FinBERT Encoder Generator & 49.53 & 52.08 \\
        \hline
        SecBERT Internal Retriever + DPR-FAISS External Retriever \newline + FinBERT Encoder Generator & 48.77 & 50.90 \\
        \bottomrule
    \end{tabular}
    \caption{Selected neural-symbolic models' performance}
    \label{tab:symbNeur}
\end{table}

\begin{table}[htbp]
    \centering
    \begin{tabular}{p{12cm} c}
        \hline
        %\textbf{Architecture} & \textbf{Dev Execution Accuracy \%} \\
        \textbf{Architecture} & \textbf{Execution }  \\ & \textbf{Accuracy \%} \\
        \hline
        Gemini-1.0-pro Zero-shot & 34.76 \\
        \hline
        Gemini-1.5-pro-latest Zero-shot & 36.27 \\
        \hline
        SecBERT Internal Retriever + Gemini-1.0-pro Generator & 39.30\\
        \hline
        SecBERT Internal Retriever + Gemini-1.5-pro-latest Generator & 41.84\\
        \hline
        SecBERT Internal Retriever + Gemini-1.0-pro Generator + Few Shot Prompt & 22.03\\
        \hline
        SecBERT Internal Retriever + Gemini-1.5-pro-latest Generator + Few Shot Prompt & 66.02\\
        \hline
        SecBERT Internal Retriever + DPR-FAISS External Retriever & 41.75\\
        + Gemini-1.0-pro Generator &  \\
        \hline
        SecBERT Internal Retriever + DPR-FAISS External Retriever & 45.33\\
        + Gemini-1.5-pro-latest Generator & \\
        \hline
        SecBERT Internal Retriever + DPR-FAISS External Retriever & 24.69\\
        + Gemini-1.0-pro Generator + Few Shot Prompt &  \\
        \hline
        \textbf{SecBERT Internal Retriever + DPR-FAISS External Retriever} & \textbf{69.37}\\
        \textbf{+ Gemini-1.5-pro-latest Generator + Few Shot Prompt} &  \\
        \hline
    \end{tabular}
    \caption{Selected prompt-based models' performance}
    \label{tab:promptBased}
\end{table}

Table~\ref{tab:symbNeur} summarizes the results for neural symbolic models (see Appendix D for full results). We observe over a 3\% performance improvement in both program accuracy and execution accuracy when we scale the generator encoder from RoBERTa Base to Large. This result, with extra model parameters, agrees with the scaling laws for Neural Language Models~\cite{ScalingLaw}. Moreover, increasing the pre-training data size seems to have a larger effect on performance than increasing model parameters. Both accuracies of the smaller RoBERTa Base model are over 2.5\% higher than those of the BERT Large model. This result supports the empirical evidence from the scaling law paper and suggests that the quality and quantity of pre-training data can be more critical than model parameter size. The larger pre-training datasets from RoBERTa Base are likely to contain more relevant information for our financial QA task, which generalizes better with higher accuracy and robustness.

Table~\ref{tab:promptBased} summarizes the results for prompt-based LLMs (see Appendix D for full results). We observe a performance improvement of around 2\% in execution accuracy when switching from Gemini-1.0-pro to the latest Gemini-1.5-pro-latest for zero-shot and single retriever only. This confirms that LLMs can perform some math calculations on their own without supervision. However, these results are worse than those from supervised neural symbolic methods. One possible reason is that without explicit supervision and structured symbolic reasoning, the LLM model might not have encountered a similar paradigm to our financial QA task setting during pre-training and does not understand the expected formats of the query and the answer in the specific finance domain. Therefore, it might not generalize well for our challenging financial numerical reasoning tasks without task-specific training.

\section{Analysis}
%Your report should include \textit{qualitative evaluation}. That is, try to understand your system (e.g., how it works, when it succeeds and when it fails) by inspecting key characteristics or outputs of your model.

\subsection{Quality of the external facts}

Even though SecBERT achieves higher average similarity scores, the retrieved external facts are less relevant to the query compared to DPR encoding. For example, when asked about the average payment volume per transaction for American Express, the SecBERT encoder achieves a 0.9359 similarity score and retrieves the definition for “zero plus tick” (a security sale term), which is entirely irrelevant to the original query. In contrast, the DPR encoder, with a 0.8413 similarity score, correctly retrieves the definition for American Express (as a financial institution). One possible explanation is that DPR is specifically trained for passage retrieval and fine-tuned with a contrastive loss for question answering tasks, optimizing it for higher relevance between financial queries and term definitions. Additionally, we randomly select 50 examples from the dev set and conduct human evaluations on the quality of retrieved external facts using SecBERT and DPR encodings. The evaluations confirm the better retrieval quality of the DPR encoding, which we choose as our optimal encoding to incorporate external knowledge for our downstream generator.

\subsection{Ablation experiments for the external retriever}

In neural symbolic models, our best model (SecBERT Internal Retriever + DPR-FAISS External Retriever + RoBERTa Large Encoder Generator) outperforms the baseline by 3.46\% in execution accuracy and 2.67\% in program accuracy, with most of the gains (around 2\%) seem to come from the stronger internal retriever SecBERT. The performance improves with better internally retrieved facts.

We generally observe slight (<1.5\%) performance improvements when adding the DPR-FAISS external retrieved facts, especially with larger encoders such as RoBERTa Large and BERT Large. The added external facts help the model better locate the correct position to extract information from the internal contexts and reinforce the key terms in the query. For example, for the dev set query “What was the change in millions of operating income from 2016 to 2017?”, without external facts, our best model only extracts the operating income in 2016 and subtracts its closest number. After adding external facts about operating incomes (how to calculate operating income), the model correctly subtracts the two operating income numbers.

However, with smaller models for the generator encoders such as RoBERTa Base, the performance remains roughly the same after including the external facts, and it even drops around 1\% for FinBERT Base with external facts. Adding external information causes about 1.4\% of the training data inputs (87/6251) to exceed the 512 max token length, resulting in truncation when fed into the model. These incomplete sentences may become irrelevant and increase hallucinations for the generator, especially for smaller models. Consequently, even with added external facts, the model might still be distracted by these incomplete sentences and generate incorrect outputs.

Therefore, we observe a trade-off between domain knowledge and hallucinations. Generally, for larger models, the gains from external facts outweigh the hallucination loss. Our ablation experiments on LLMs also confirm this hypothesis, showing solid improvements of over 2\% when adding external facts. However, for smaller models the hallucination loss starts to increase and may even offset the external knowledge gain. This interesting finding highlights the complex dynamics between model size, external knowledge integration, and hallucination effects for future study.

\subsection{Ablation experiments for the few-shot prompts in LLM}

When we add few-shot prompts to Gemini-1.0-pro, we surprisingly see a large performance drop from 39.30\% to 22.03\%. Analyzing the generated outputs, it seems the LLM sometimes uses numbers from the contexts of few-shot examples to answer the new query. For example, fo the question “what percentage of total purchase commitments are due after 2014?”, Gemini-1.0-pro generates the incorrect output 49.49\% by directly using the numerical answer 0.4949 from the first example in few-shot prompts, which is entirely unrelated to the target query. 
%These added few-shot examples increase the hallucination loss for the model and lead to worse performance.

Moreover, the LLM also tends to ignore the numbers in the given query contexts and can disregard the actual context. For example, even with an extremely short query context such as “2012: 720; thereafter: 4717; total debt: \$7680; what percentage of total debt is due after 2012?”, Gemini-1.0-pro still provides the incorrect answer 84.37\%, whereas the correct answer is 61.42\% (4717/7680). Gemini-1.5-pro-latest also gives the wrong answer 81.76\%. Without additional instructions, the LLM may prefer to use the knowledge acquired during pre-training rather than the actual context for QA.

However, after applying the same few-shot prompts to Gemini-1.5-pro-latest, we find it can answer the above question perfectly. By adding the few-shot prompts, the execution accuracy greatly improves to 66.02\%, and we achieve our SOTA model by adding external facts with 69.37\%. These results suggest that Gemini-1.5-pro-latest may have significantly improved capabilities in numerical reasoning and arithmetic operations, especially with better optimized few-shot learning capabilities. One possible explanation is that Gemini-1.5-pro-latest includes advancements that specifically address the issue of hallucination, where the model generates information not grounded in the input context. This helps the model rely more on the provided context and less on unrelated pre-training knowledge.

\subsection{Performance breakdown across sub-datasets}

\begin{table}[htbp]
    \centering
    \begin{tabular}{lcc}
        \toprule
        \textbf{Sub Dataset} & \textbf{Best Neural Symbolic} & \textbf{Best Prompt-based LLM} \\
         & \textbf{Execution Accuracy \%} & \textbf{Execution Accuracy \%} \\
        \midrule
        Full Test Set & 61.28 & 68.39 \\
        Table-only Questions & 68.55 & 71.87 \\
        Text-only Questions & 56.13 & 67.41 \\
        Table-Text-mixed Questions & 44.26 & 64.72 \\
        Long Context Questions & 59.35 & 67.24 \\
        Short Context Questions & 62.11 & 68.70 \\
        Single Reasoning Step Questions & 69.97 & 75.48 \\
        Multi Reasoning Step Questions & 46.60 & 58.62 \\
        \bottomrule
    \end{tabular}
    \caption{Performance of models on different types of questions sub-datasets}
    \label{tab:sub_datasets}
\end{table}

Table~\ref{tab:sub_datasets} shows the performance breakdown of our best models across sub test sets for various types of questions. With only 20 epochs of training, our best neural symbolic model even slightly outperforms the best model from the original FinQA paper (61.24\%) that had 300 epochs of training in terms of execution accuracy. Our best LLM model achieves the SOTA with 68.39\% execution accuracy. We observe that both selected models perform better on table-only questions. 
%With more unified structures and shorter contexts, these questions are less challenging than table-text-mixed questions.

Both models perform slightly worse on originally long context questions with more irrelevant content. This robustness could further confirm the effectiveness of using internal retrievers to filter out unimportant sentences, ensuring all model input texts are short enough regardless of the original context length. It seems that originally long questions are more challenging than short ones.

Most questions require single-step reasoning, and both models perform much better on these easier questions. Not surprisingly, they both perform much worse on multi-step reasoning questions compared to their accuracy on the full set. Overall, the best LLM model achieves better results on all subsets compared to the best neural symbolic model. This further confirms the strong numerical reasoning ability of Gemini-1.5-pro-latest in domain-specific QA tasks.

\subsection{Error Analysis}

We randomly select 30 error cases from our best model results. Our analysis indicates that over 65\% of the errors are due to poor numerical reasoning for numerical unit conversions and incorrect number retrieval, especially for multi-step questions. For example, a question asks for the net sales in billions, but our model predicts 7,222.22 (the correct answer is 7.22). See Appendix E for more details.

\section{Conclusion \& Future Work}
This project implements a multi-retriever RAG system and leverages the latest LLM to address numerical reasoning QA in the finance domain. We conduct comprehensive ablation experiments and error analysis across various models. Largely due to the domain-specific training of the SecBERT encoder, the best neural symbolic model outperforms our baseline from the best models in the FinQA paper, and the best prompt-based LLM generator achieves SOTA with significant improvements (>7\%). The results demonstrate the trade-off effects of hallucinations for added external knowledge on smaller models and extra few-shot examples. Our analysis further confirms the improved numerical reasoning ability of the latest LLM, which is better optimized for few-shot learning.

\begin{figure}[h]
    \begin{center}
        \includegraphics[width=0.85\textwidth]{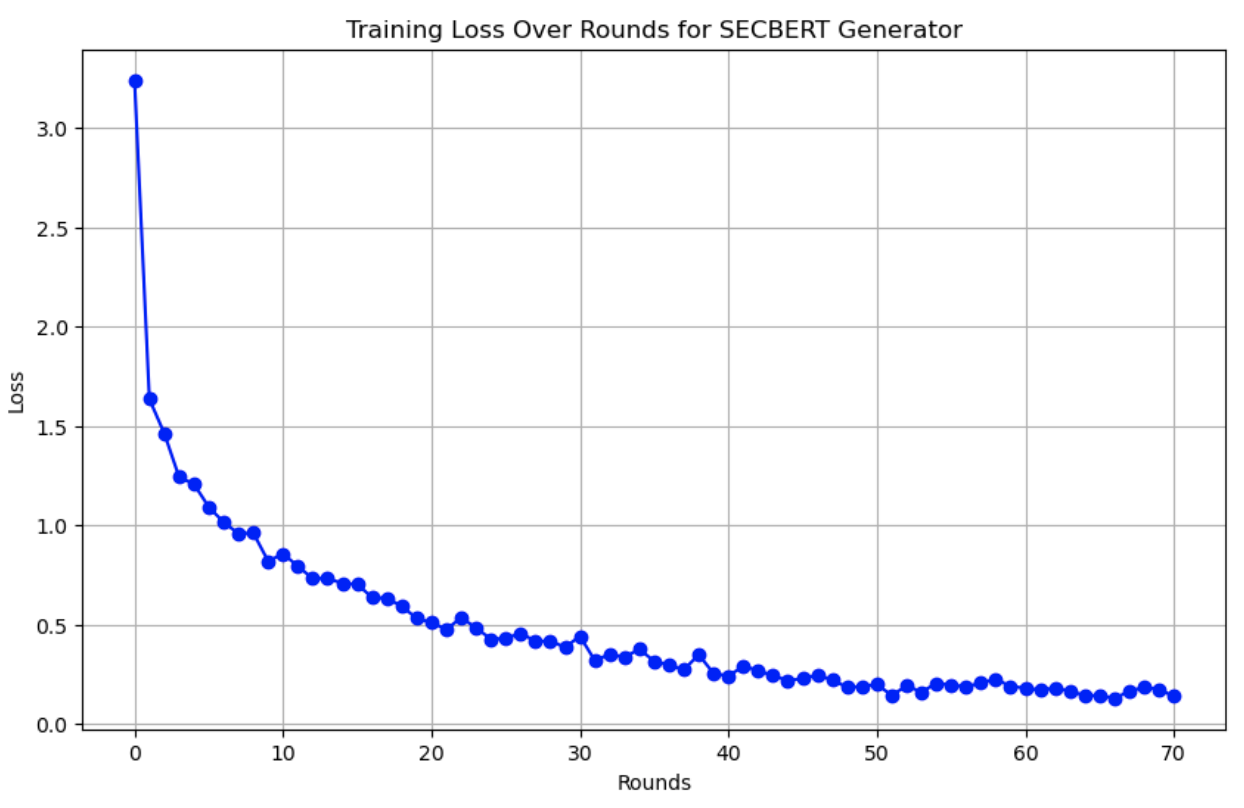}
        \caption{Training Loss Over Rounds (20 epochs) for the Generator with SEC-BERT encoder and LSTM decoder.}
        \label{fig:sec-bert-train-loss-generator}
    \end{center}
\end{figure}

Computing power was a significant limitation for our project. Despite training each model for only 20 epochs, compared to over 300 epochs by the FinQA authors, our best symbolic neural model still outperforms their best model. Figure~\ref{fig:sec-bert-train-loss-generator} shows that the loss continues to decrease after 20 epochs, suggesting that additional future training could further improve performance. 

Another limitation we identified was the length of the context we could retrieve. Some added contexts are truncated due to max token length limits, which could reinforce the hallucination effects. In the future, we could annotate some data and fine-tune an NER system to identify key financial terms for each query for better and shorter external knowledge retrieval.

\bibliographystyle{unsrt}
\bibliography{references}

\newpage

\section{Appendices}

{\fontsize{11}{12}\selectfont \textbf{A: Prompts Used for the LLM-based Approach: }}

\textbf{The instruction prompt at the end of the input:  } “given the contexts, your generated response to the above question must only be a single numerical number only (without any symbols nor texts), or a numerical number with a percentage sign only, or just yes/no, whichever applicable.”

\textbf{The first few-shot example prompt at the beginning of the input:  } "as of and for the years ended december 31 , the operating income of 2003 is 1039 ; the operating income of 2002 ( 1 ) is 695 ; the operating income of 2001 ( 1 ) is 1717 ; what was the percentage change in operating income for entities in which the company has the ability to exercise significant influence but does not control and that are accounted for using the equity method between 2002 and 2003? given the contexts, your generated response to the above question must only be a single numerical number only (without any symbols nor texts), or a numerical number with a percentage sign only, or just yes/no, whichever applicable. 0.4949.”

\textbf{The second few-shot example prompt at the beginning of the input:  } "in millions except for per share data the diluted-as reported of 2005 is \$ 4.55 ; in millions except for per share data the diluted-pro forma of 2005 is 4.52 ; was diluted-as reported net income per share greater than diluted-pro forma net income per share? given the contexts, your generated response to the above question must only be a single numerical number only (without any symbols nor texts), or a numerical number with a percentage sign only, or just yes/no, whichever applicable. Yes."
\\ 

{\fontsize{11}{12}\selectfont \textbf{B: Definitions for Special Tokens}}

\textbf{Operation tokens: } 'EOF', 'UNK', 'GO', ')', 'add(', 'subtract(', 'multiply(', 'divide(', 'exp(', 'greater(', 'table\_sum(', 'table\_average(', 'table\_max(', 'table\_min('.

\textbf{Step memory tokens: } \#0, \#1, \#2, \#3, \#4, \#5, \#6, \#7, \#8, \#9, \#10.

\textbf{Constant tokens: } CONST\_2, CONST\_1, CONST\_3, CONST\_4, CONST\_5, CONST\_6, CONST\_7, CONST\_8, CONST\_9, CONST\_10, CONST\_100, CONST\_1000, CONST\_10000, CONST\_100000, CONST\_1000000, CONST\_10000000, CONST\_1000000000, CONST\_M1.
\\

{\fontsize{11}{12}\selectfont \textbf{C: Training Loss Trend }}

{\fontsize{11}{12}\selectfont \textbf{D: Full Results Table }}

We split the table with all of our full results into two so it can fit into one page:

\begin{table}[htbp]
    \centering
     \scalebox{0.9}{%
    \begin{tabular}{lcc}
        \toprule
        \textbf{Architecture (20 epochs of training)} & \textbf{Dev Program Accuracy \%} & \textbf{Dev Execution Accuracy \%} \\
        \midrule
        BERT Base Internal Retriever ONLY & & \\
        + RoBERTa Large Encoder Generator (Baseline) & 57.87 & 60.02 \\
        \midrule
        BERT Base Internal Retriever & & \\
        + DPR-FAISS External Retriever & & \\
        + RoBERTa Large Encoder Generator & 58.49 & 60.96 \\
        \midrule
        SecBERT Internal Retriever ONLY & & \\
        + RoBERTa Large Encoder Generator & 59.70 & 62.11 \\
        \midrule
        SecBERT Internal Retriever & & \\
        + DPR-FAISS External Retriever & & \\
        + RoBERTa Large Encoder Generator & 60.54 & 63.48 \\
        \midrule
        SecBERT Internal Retriever ONLY & & \\
        + BERT Large Encoder Generator & 53.95 & 55.76 \\
        \midrule
        SecBERT Internal Retriever & & \\
        + DPR-FAISS External Retriever & & \\
        + BERT Large Encoder Generator & 54.64 & 56.97 \\
        \midrule
        SecBERT Internal Retriever ONLY & & \\
        + SecBERT Encoder Generator & 51.76 & 54.36 \\
        \midrule
        SecBERT Internal Retriever & & \\
        + DPR-FAISS External Retriever & & \\
        + SecBERT Encoder Generator & 51.84 & 54.52 \\
        \midrule
        SecBERT Internal Retriever ONLY & & \\
        + BERT Base Encoder Generator & 49.78 & 52.61 \\
        \midrule
        SecBERT Internal Retriever & & \\
        + DPR-FAISS External Retriever & & \\
        + BERT Base Encoder Generator & 49.69 & 52.63 \\
        \midrule
        SecBERT Internal Retriever ONLY & & \\
        + FinBERT Encoder Generator & 49.53 & 52.08 \\
        \midrule
        SecBERT Internal Retriever & & \\
        + DPR-FAISS External Retriever & & \\
        + FinBERT Encoder Generator & 48.77 & 50.90 \\
        \midrule
        SecBERT Internal Retriever ONLY & & \\
        + RoBERTa Base Encoder Generator & 56.49 & 58.33 \\
        \midrule
        SecBERT Internal Retriever & & \\
        + DPR-FAISS External Retriever & & \\
        + RoBERTa Base Encoder Generator & 56.75 & 58.81 \\
        \midrule
        BERT Base Internal Retriever ONLY & & \\
        + FinBERTQA Encoder Generator & 40.53 & 44.22 \\
        \bottomrule
    \end{tabular}}
    \caption{Results table (Part 1)}
    \label{tab:model_architectures_part1}
\end{table}

\begin{table}[htbp]
    \centering
     \scalebox{0.9}{%
    \begin{tabular}{lcc}
        \toprule
        \textbf{Architecture (20 epochs of training)} & \textbf{Dev Program Accuracy \%} & \textbf{Dev Execution Accuracy \%} \\
        \midrule
        SpanBERT Internal Retriever ONLY & & \\
        + FinBERT Encoder Generator & 45.23 & 44.12 \\
        \midrule
        SpanBERT Internal Retriever ONLY & & \\
        + SecBERT Encoder Generator & 44.03 & 42.45 \\
        \midrule
        SpanBERT Internal Retriever ONLY & & \\
        + BERT Base Encoder Generator & 42.11 & 43.68 \\
        \midrule
        Gemini-1.0-pro Zero-shot & N/A & 34.76 \\
        \midrule
        Gemini-1.5-pro-latest Zero-shot & N/A & 36.27 \\
        \midrule
        BERT Base Internal Retriever ONLY & N/A & 35.56 \\
        + Gemini-1.0-pro Generator & & \\
        \midrule
        BERT Base Internal Retriever & N/A & 36.35 \\
        + DPR-FAISS External Retriever &  & \\
        + Gemini-1.0-pro Generator &  &  \\
        \midrule
        BERT Base Internal Retriever & N/A & 21.32\\
        + DPR-FAISS External Retriever &  & \\
        + Gemini-1.0-pro Generator &  & \\
        + Few shot Prompt &  &  \\
        \midrule
        SecBERT Internal Retriever ONLY & N/A & 39.30 \\
        + Gemini-1.0-pro Generator &  &  \\
        \midrule
        SecBERT Internal Retriever ONLY & N/A & 41.84 \\
        + Gemini-1.5-pro-latest Generator &  &  \\
        \midrule
        SecBERT Internal Retriever ONLY & N/A & 22.03 \\
        + Gemini-1.0-pro Generator & & \\
        + Few Shot Prompt & &  \\
        \midrule
        SecBERT Internal Retriever ONLY & N/A &  66.02\\
        + Gemini-1.5-pro-latest Generator &  & \\
        + Few Shot Prompt & & \\
        \midrule
        SecBERT Internal Retriever & N/A & 41.75 \\
        + DPR-FAISS External Retriever & & \\
        + Gemini-1.0-pro Generator & &  \\
        \midrule
        SecBERT Internal Retriever & N/A & 45.33 \\
        + DPR-FAISS External Retriever &  & \\
        + Gemini-1.5-pro-latest Generator &  &  \\
        \midrule
        SecBERT Internal Retriever & N/A & 24.69 \\
        + DPR-FAISS External Retriever &  & \\
        + Gemini-1.0-pro Generator & & \\
        + Few Shot Prompt &  &  \\
        \midrule
        SecBERT Internal Retriever & N/A & 69.37 \\
        + DPR-FAISS External Retriever &  & \\
        + Gemini-1.5-pro-latest Generator &  & \\
        + Few Shot Prompt &  & \\
        \midrule
        SecBERT Internal Retriever & N/A & 0 \\
        + DPR-FAISS External Retriever &  & \\
        + BERT encoder + gpt2 decoder &  &  \\
        \midrule
        T5-Base Generator (zero shot) & N/A & 2.83 \\
        \midrule
        BERT Base Internal Retriever & N/A & 2.15 \\
        + T5-Base Generator (zero shot) &  &  \\
        \midrule
        BERT Base Internal Retriever & N/A & 2.15 \\
        + DPR-FAISS External Retriever &  & \\
        + T5-Base Generator &  & \\
        \bottomrule
    \end{tabular}}
    \caption{Results table (Part 2)}
    \label{tab:model_architectures_part2}
\end{table}

\newpage 
{\fontsize{11}{12}\selectfont \textbf{E: Error Cases}}

\textbf{Case One: numerical unit conversions. } The question asks "what is the applied 2019s net sales in 2018, (in billions)?". The correct reasoning steps are: divide(1.3, divide(18, 100)) = 7.22, but the model answers 7,222.22 without properly identifying the requirements for the question and converting the number in billions.  

\textbf{Case Two: number retrieval. } The question asks "what is the average amortization amount, in millions, from 2015-2019?". The correct reasoning steps are: "divide(add(multiply(45, 4), 44), 5)" = 44.8, but the model incorrectly retrieves the 36 million amortization expense ending in December 31 , 2014 instead of 45 million for 2015.

\end{document}